\documentclass[sigconf]{acmart}

\usepackage{algorithm}
\usepackage{algorithmic}
\usepackage{mathtools}
\usepackage{amsfonts}
\usepackage{amsmath}
\usepackage{bbm}
\usepackage{comment}
\usepackage{color}
\usepackage{bm}
\usepackage{url}
\usepackage{enumitem}
\usepackage{booktabs}
\usepackage{multirow}
\usepackage{color,soul}
\usepackage{array}
\usepackage{subcaption,graphicx}
\usepackage{balance}

\newcommand{\model}{{SPAC}}

\AtBeginDocument{%
  \providecommand\BibTeX{{%
    \normalfont B\kern-0.5em{\scshape i\kern-0.25em b}\kern-0.8em\TeX}}}

\copyrightyear{2022} 
\acmYear{2022} 
\setcopyright{rightsretained} 
\acmConference[KDD '22]{Proceedings of the 28th ACM SIGKDD Conference on Knowledge Discovery and Data Mining}{August 14--18, 2022}{Washington, DC, USA}
\acmBooktitle{Proceedings of the 28th ACM SIGKDD Conference on Knowledge Discovery and Data Mining (KDD '22), August 14--18, 2022, Washington, DC, USA}
\acmDOI{10.1145/3534678.3539435}
\acmISBN{978-1-4503-9385-0/22/08}

\usepackage{etoolbox}
\makeatletter
\patchcmd{\maketitle}{\@copyrightpermission}{
\begin{minipage}{0.3\columnwidth}
\href{https://creativecommons.org/licenses/by/4.0/}{\includegraphics[width=0.90\textwidth]{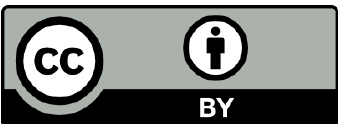}}
\end{minipage}\hfill
\begin{minipage}{0.7\columnwidth}
\href{https://creativecommons.org/licenses/by/4.0/}{This work is licensed under a Creative Commons Attribution International 4.0 License.}
\end{minipage}

\vspace{5pt}
}{}{}

\makeatother

\begin{document}

\title{Graph Structural Attack by Perturbing Spectral Distance}

\author{Lu Lin}
\email{ll5fy@virginia.edu}
\affiliation{
    \institution{University of Virginia} 
    \city{Charlottesville}
    \state{VA 22904}
    \country{USA}
 }
 
\author{Ethan Blaser}
\email{ehb2bf@virginia.edu}
\affiliation{
    \institution{University of Virginia} 
    \city{Charlottesville}
    \state{VA 22904}
    \country{USA}
 }
 
\author{Hongning Wang}
\email{hw5x@virginia.edu}
\affiliation{
    \institution{University of Virginia} 
    \city{Charlottesville}
    \state{VA 22904}
    \country{USA}
 }

\renewcommand{\shortauthors}{Lu Lin, Ethan Blaser, \& Hongning Wang}

\begin{abstract}
Graph Convolutional Networks (GCNs) have fueled a surge of research interest due to their encouraging performance on graph learning tasks, but they are also shown vulnerability to adversarial attacks. 
In this paper, an effective graph structural attack is investigated to disrupt graph spectral filters in the Fourier domain, which are the theoretical foundation of GCNs.
We define the notion of spectral distance based on the eigenvalues of graph Laplacian to measure the disruption of spectral filters.
We realize the attack by maximizing the spectral distance and propose an efficient approximation to reduce the time complexity brought by eigen-decomposition.
The experiments demonstrate the remarkable effectiveness of the proposed attack in both black-box and white-box settings for both test-time evasion attacks and training-time poisoning attacks.
Our qualitative analysis suggests the connection between the imposed spectral changes in the Fourier domain and the attack behavior in the spatial domain, which provides empirical evidence that maximizing spectral distance is an effective way to change the graph structural property and thus disturb the frequency components for graph filters to affect the learning of GCNs.
\end{abstract}



\maketitle

\section{Introduction}
Graph signal processing applies the idea of signal processing to graph data, allowing existing signal processing tools such as spectral filtering and sampling to be used for learning graph embeddings \cite{dong2020graph}.
In particular, spectral filters are generalized to create Graph Convolutional Networks (GCNs), which have prominently advanced the state of the art on many graph learning tasks \cite{kipf2016semi, velivckovic2017graph}.
However, despite their great success, recent works show that GCNs exhibit vulnerability to \textit{adversarial perturbations}: such models can be easily fooled by small perturbations on graph structure or node properties, and thus generate inaccurate embeddings leading to erroneous predictions in downstream tasks \cite{zugner2018adversarial, dai2018adversarial, zugner2019adversarial, xu2019topology, wu2019adversarial}.

Graph data differs from image or text data due to the topological structure formed among nodes.  
On one hand, GCNs exploit such structures to aggregate information conveyed in nodes' neighborhoods, which yield better predictive power on many tasks (such as link prediction \cite{schlichtkrull2018modeling} and node classification \cite{rong2019dropedge}). 
But on the other hand, the complex dependency relations introduced by the topological structure of graphs also expose learning models to a greater risk: an attacker can mislead classifiers to erroneous predictions by just slightly perturbing the graph structure, without even modifying any node features. 
Various adversarial attacks have been studied on graph structure \cite{jin2020adversarial}, considering different prediction tasks (node classification \cite{xu2019topology, wang2019attacking, zugner2019adversarial} or graph classification \cite{dai2018adversarial, ma2019attacking}), attacker's knowledge (white-box \cite{xu2019topology, wang2019attacking, zugner2018adversarial} or black-box \cite{dai2018adversarial, ma2019attacking}), attack phases (test-time evasion attack \cite{chang2020restricted, dai2018adversarial, xu2019topology} or training-time poisoning attack \cite{ wang2019attacking, zugner2019adversarial}), and perturbation types (edge modification \cite{xu2019topology, zugner2019adversarial} or node modification \cite{sun2019node}).
In this paper, we focus on the structural attack by adding or removing edges to compromise the node classification performance of a victim GCN model. 

Graph convolution, as the fundamental building block of GCNs, is designed to filter graph signals in the \textit{Fourier} domain. Studies in spectral graph theory \cite{chung1997spectral} show that the spectra (eigenvalues) of the graph Laplacian matrix capture graph structural properties (e.g., the second smallest eigenvalue, also known as the Fiedler value, reflects the algebraic connectivity of graphs \cite{mohar1991laplacian}
). 
Therefore exploiting spectral changes provides a comprehensive way to study the vulnerability of GCN models. However, so far most structural attack solutions only search for perturbations in the \textit{spatial} domain.
Ignoring the direct source of GCN models' vulnerability which resides in the Fourier domain limits the effectiveness of attacks. 

\begin{figure}[t]
 \centering
  \includegraphics[width=0.45\textwidth]{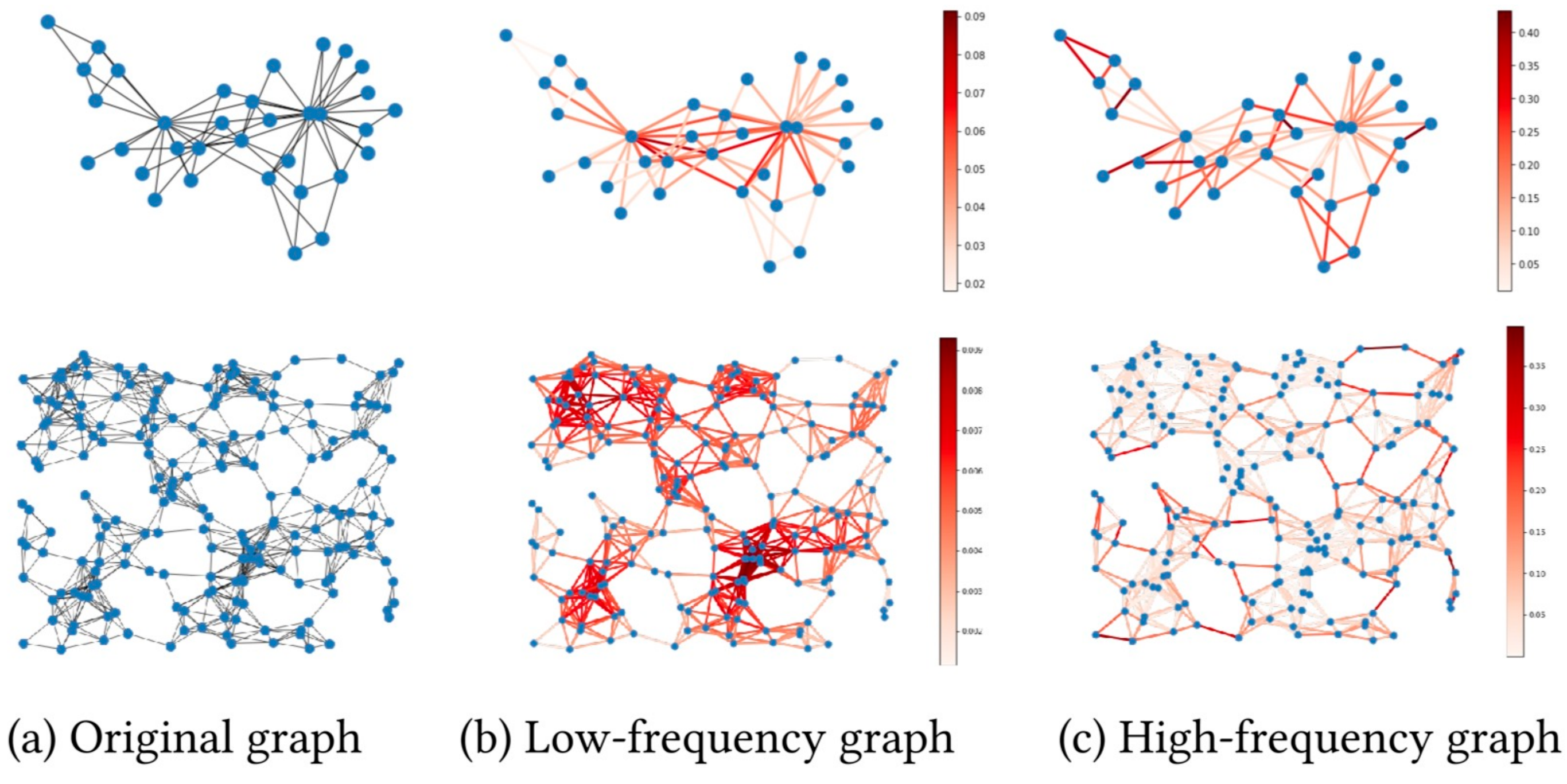}
  \vspace{-1mm}
 \caption{Relationship between graph structural property and frequency components in Fourier domain on the Karate club graph (TOP) and a random geometric graph (BOTTOM). Color denotes the edge reconstructed when only using low-frequency components (b) or high-frequency components (c). Darker color indicates a larger edge reconstruction value.}
 \label{fig:example}
 \vspace{-3mm}
\end{figure}

Studying GCN models' vulnerability in the Fourier domain can effectively capture important edges that influence the structural property the most, e.g., the clustering structure of nodes. 
According to the concept of graph signal processing, the eigen-decomposition of the Laplacian matrix of a graph defines the frequency domain of message passing on the graph.
Recent works have established the relationship between frequency components and graph clustering \cite{donnat2018learning, song2018constructing}. 
Based on the ascending ordered eigenvalues of the Laplacian matrix,
we can obtain both low- and high-frequency components, which play different roles in message passing on graphs.
The eigenvectors associated with small eigenvalues carry smoothly varying signals, encouraging neighbor nodes to share similar properties (e.g., nodes within a cluster).
In contrast, the eigenvectors associated with large eigenvalues carry sharply varying signals across edges (e.g., nodes from different clusters) \cite{chang2021spectral, jin2019power}.
Figure \ref{fig:example} illustrates the concept on both the popularly studied social network graph Karate club and a random geometric graph. 
We visualize the edges by their reconstruction using the eigenvectors only associated with the top low- or high-frequency components in each graph respectively. For example, in Figure \ref{fig:example} (b), the color of each edge reflects its reconstruction only using eigenvectors associated with the lowest eigenvalues.
On one hand, the information inside the closely connected components is retained when low-frequency components are used to reconstruct the graph; and on the other hand, the inter-cluster information is captured when constructing the graph with only high-frequency components. 
This example clearly illustrates that graph frequency components encode the global structure of graphs, which motivates us to study GCN models' vulnerability in the Fourier domain.

In this work, we propose a principled graph perturbation strategy in the Fourier domain to improve the effectiveness of adversarial attacks against GCN models.
Specifically, we define the \textit{spectral distance} between the original and perturbed graph, measured by the change in their Laplacian eigenvalues. 
Then we build a structural attack model which directly maximizes the spectral distance in a black-box fashion. 
To solve this combinatorial optimization problem, we relax the binary constraint on edge perturbation to a continuous one, and apply a randomization sampling strategy to generate valid binary edge perturbations.  
We name this method \textbf{SP}ectral \textbf{A}tta\textbf{C}k, abbreviated as \textbf{\model{}}.
It is worth noting that generating the \model{} attack requires eigen-decomposition of the Laplacian matrix, which results in a time complexity of $\mathcal{O}(n^3)$ with $n$ nodes in a graph. To handle large graphs, we propose an approximation solution only based on a set of largest and smallest eigenvalues and their corresponding eigenvectors, and use eigenvalue perturbation theory \cite{stewart1990matrix} to avoid frequent computation of eigen-decomposition, which reduces the time complexity to $\mathcal{O}(n^2)$. 
Our attack method is evaluated under both white-box and black-box settings for both evasion and poisoning attacks on a set of benchmark graph datasets.
Promising empirical results demonstrate that convolutional graph learning models are sensitive to spectral changes, which expands the scope of adversarial attacks on graphs to the Fourier domain and opens up new possibilities to verify and enhance GCNs' robustness in both the spatial and Fourier domains.

\section{Related Work}
Adversarial attacks on graph structures have been extensively studied in recent years.
The vast majority of attack efforts manipulate graphs in the spatial domain to maximize a task-specific attack objective.
However, the vulnerability of graph convolutions in the Fourier domain is less studied in existing attack solutions.
We bridge the gap by measuring and maximizing the spectral changes in the graph Laplacian matrix, such that we can directly disrupt the graph spectral filters and attack the graph convolutions.

\noindent\textbf{Adversarial attack on graph structures.}
The attacker aims to perturb the graph adjacency matrix in directions that lead to large classification loss. In the white-box setting, the attacker follows the gradients on the adjacency matrix to find such perturbations \citep{zugner2019adversarial, xu2019topology, wu2019adversarial, zugner2018adversarial, chen2018fast, li2021adversarial, wang2020evasion}.
Different strategies are exploited to convert continuous gradients into binary edge modifications.
Topology attack \citep{madry2017towards} uses randomization sampling to select sub-optimal binary perturbations.
Nettack \citep{zugner2018adversarial} and FGA \citep{chen2018fast} select edge changes with the largest gradient greedily. 
Metattack \citep{zugner2019adversarial} first calculates meta-gradient on graph adjacency matrix to solve a bi-level optimization problem for poisoning attack, and then greedily picks perturbations with the largest meta-gradient. 
In the black-box setting, the attacker cannot access gradients of the victim model but uses a proxy (e.g. model output scores) to search for the best perturbations \citep{ma2020towards, ma2019attacking, dai2018adversarial}.
Reinforcement learning based solutions  \citep{chang2020restricted, ma2019attacking, dai2018adversarial, raman2020learning} make a series of edge addition or deletion decisions that yield the maximal attack utility and thus can serve for black-box setting. 


These attacks search for perturbations in the spatial space, but the target GCNs generate node embeddings by the signal filters defined in the Fourier space. Thus the vulnerability of graph convolutions reflected on the graph spectral changes cannot be fully realized. Our method captures such vulnerability directly in the Fourier domain measured by the spectral distance between the original and perturbed graphs for a more effective attack.

\noindent\textbf{Spectral perturbations on graphs.}
Existing attack methods in the Fourier space are generally sparse.
\citet{bojchevski2019adversarial} reformulate random walk based models as a matrix factorization problem, and propose an attack strategy to search for edges that lead to large eigenvalue changes in the derived matrix. However, this method is model-specific and cannot be easily applied to general forms of GCNs.
GF-Attack \cite{chang2020restricted} constructs an objective based on the low-rank graph spectra and feature matrix to guide the attack in a black-box fashion.
A universal attack on deformable 3D shape data is proposed to change the scale of its eigenvalues \cite{rampini2021universal}, but it is not studied in the graph domain. 
DICE \citep{waniek2018hiding} corrupts the graph structure by ``deleting edges internally and connecting nodes externally'' across clusters which implicitly influences the graph's spectral property. 
But this heuristic is performed without any principled guidance. 
Studies that analyze spectral graph filters \cite{kenlay2021interpretable, levie2019transferability, kenlay2020stability} provide the theoretical stability upper bounds of popular filters used in graph convolution models, such as polynomial and identity filters. It is shown that the filters become unstable if the end nodes of changed edges have low degrees or the perturbation is concentrated spatially around any single node \cite{kenlay2021interpretable}. 
Our method empirically shows that we can attack the vulnerability of these filters and break such requirements by directly maximizing graph spectral changes.
\section{Spectral Attack on Graphs}
In this section, we first briefly discuss the spectral graph filters which are the key building blocks of graph convolution models. 
We then propose to maximize the changes on the graph Laplacian spectrum, such that we can exploit the edge perturbation budget to most effectively influence the spectral filters and attack graph convolutions. 
We solve the resulting optimization problem using gradient descent, and propose an efficient approximation via eigenvalue perturbation theory on selective eigenvalues. 
We finally discuss the theoretical evidence showing the dependency between the eigenvalues of graph Laplacian and the stability of GCN models, which supports the proposed spectral attack on graph data.

\subsection{Preliminaries}
\noindent\textbf{Notations.}
Let $G=(V, E)$ be a connected undirected graph with $n$ nodes and $m$ edges. Let $\mathbf{A}\in\{0, 1\}^{n\times n}$ be its adjacency matrix. The diagonal degree matrix can be calculated by $\mathbf{D}=\text{diag}(\mathbf{A}\mathbf{1}_n)$ with entry $\mathbf{D}_{ii}=d_i=\sum^{n}_{j=1}\mathbf{A}_{ij}$, and $\mathbf{1}_n$ is an all-one vector with dimension $n$. 
The normalized Laplacian matrix of a graph is defined as $\mathbf{L}=\text{Laplacian}(\mathbf{A})=\mathbf{I}_n-\mathbf{D}^{-1/2}\mathbf{A}\mathbf{D}^{-1/2}$, where $\mathbf{I}_n$ is an ${n\times n}$ identity matrix.
Since $\mathbf{L}$ is symmetric positive semi-definite, it admits an eigen-decomposition $\mathbf{L}=\mathbf{U\Lambda U^{\top}}$.
The diagonal matrix $\mathbf{\Lambda}=\text{eig}(\mathbf{L})=\text{diag}(\lambda_1, \dots, \lambda_{n})$ consists of the real eigenvalues of $\mathbf{L}$ in an increasing order such that $0=\lambda_1\leq \dots\leq\lambda_n\leq 2$, and the corresponding $\mathbf{U}=[\mathbf{u}_1,\dots,\mathbf{u}_n ]\in\mathbb{R}^{n\times n}$ is a unitary matrix where the columns consist of the eigenvectors of $\mathbf{L}$.
$\mathbf{X}\in\mathbf{R}^{n\times d}$ denotes the node feature matrix where each node $v$ is associated with a $d$-dimensional feature vector.

\noindent\textbf{Graph Fourier transform.}   
By viewing graph embedding models from a signal processing perspective, the normalized Laplacian $\mathbf{L}$ serves as a shift operator and defines the frequency domain of a graph \cite{sandryhaila2013discrete}. As a result, the eigenvectors of $\mathbf{L}$ can be considered as the graph Fourier bases, and the eigenvalues correspond to frequency components.
Take one column of $\mathbf{X}$ as an example of graph signal, which can be compactly represented as $\mathbf{x}\in\mathbb{R}^{n}$. The graph Fourier transform of graph signal $\mathbf{x}$ is given by $\hat{\mathbf{x}}=\mathbf{U}^{\top}\mathbf{x}$ and the inverse graph Fourier transform is then $\mathbf{x}=\mathbf{U}\hat{\mathbf{x}}$.
The graph signals in the Fourier domain are filtered by amplifying or attenuating the frequency components $\mathbf{\Lambda}$.  

\noindent\textbf{Spectral graph convolution.} 
At the essence of different graph convolutional models is the spectral convolution, which is defined as the multiplication of signal $\mathbf{x}$ with a filter $g_{\phi}$ parameterized by $\phi\in\mathbb{R}^n$ in the Fourier domain \cite{defferrard2016convolutional}:
\begin{align}
\label{eq:conv}
    g_{\phi}(\mathbf{L})\star \mathbf{x} &= \mathbf{U}g^*_{\phi}(\mathbf{\Lambda})\mathbf{U}^{\top}\mathbf{x}
\end{align}
where the parameter $\phi$ is a vector of spectral filter coefficients.
The filter $g_{\phi}$ defines a smooth transformation function, and a commonly used filter is the polynomial filter:
\begin{equation}
\label{eq:filter}
    g^*_{\phi}(\mathbf{\Lambda})=\sum\nolimits^{\infty}_{k=0}\phi_{k}\mathbf{\Lambda}^k
\end{equation}
which can be approximated by a truncated expansion.
A commonly adopted approximation is based on the Chebyshev polynomials $T_k(\mathbf{\Lambda})$, which are recursively defined as $T_0(\mathbf{\Lambda})=\mathbf{I}_n, T_1(\mathbf{\Lambda})=\mathbf{\Lambda}$ and $T_{k+1}(\mathbf{\Lambda})=2\mathbf{\Lambda}T_k(\mathbf{\Lambda})-T_{k-1}(\mathbf{\Lambda})$
. Using the Chebyshev polynomials up to the $K$-th order achieves the following approximation \cite{hammond2011wavelets}:
\begin{equation}
\label{eq:chev}
    g^*_{\phi}(\mathbf{\Lambda})\approx\sum\nolimits^{K}_{k=0}\phi_k T_k(\tilde{\mathbf{\Lambda}})
\end{equation}
with a rescaled $\tilde{\mathbf{\Lambda}}=2\mathbf{\Lambda}/\lambda_{n}-\mathbf{I}_n$.

\noindent\textbf{Graph Convolutional Network (GCN).} 
A vanilla GCN is a first-order approximation of the spectral graph convolution with the Chebyshev polynomials \cite{kipf2016semi}. Setting $K=1$, $\phi_0=-\phi_1$ in Eq. (\ref{eq:chev}) and approximating $\lambda_{n}\approx2$, we obtain the convolution operation $g_{\phi}(\mathbf{L})\star\mathbf{x}=(\mathbf{I}_n+\mathbf{D}^{-1/2}\mathbf{A}\mathbf{D}^{-1/2})\mathbf{x}$. We can replace matrix $\mathbf{I}_n+\mathbf{D}^{-1/2}\mathbf{A}\mathbf{D}^{-1/2}$ with a self-loop enhanced version $\tilde{\mathbf{L}}=\mathbf{\tilde{D}}^{-1/2}\mathbf{\tilde{A}}\mathbf{\tilde{D}}^{-1/2}$ where $\mathbf{\tilde{A}}=\mathbf{A}+\mathbf{I}_n$ and $\mathbf{\tilde{D}}=\mathbf{D}+\mathbf{I}_n$. This resembles the vanilla GCN layer with activation function $\sigma$ and trainable network parameters $\mathbf{\Theta}$ for feature transformation:
\begin{equation}
\label{eq:gcn}
    \mathbf{H}^{(l+1)}=\sigma\Big(\mathbf{\tilde{L}}\mathbf{H}^{(l)}\mathbf{\Theta}^{(l)}\Big)
\end{equation}
where the signals from the previous layer $\mathbf{H}^{(l)}$ is filtered to generate new signals $\mathbf{H}^{(l+1)}$. To unify the notations, $\mathbf{H}^{(0)}=\mathbf{X}$ denotes the input node features, and $\mathbf{Z}=\mathbf{H}^{(L)}$ denotes the output node embeddings of an $L$-layer GCN model.

\subsection{Spectral Distance on Graphs}
Based on the aforementioned spectral perspective for understanding GCNs, we aim to generate edge perturbations that can disrupt the spectral filters the most when processing input signals on the graph.
We measure the disruption by the changes in the eigenvalues of graph Laplacian, which we define as the spectral distance.

As shown in Eq. (\ref{eq:conv}), the spectral filters $g^*_{\phi}(\mathbf{\Lambda})$ are the key in graph convolutions to encode graph signals that are transformed in the Fourier domain. The output of the spectral filters is then transformed back to the spatial domain to generate node embeddings.
Therefore, perturbing the spectral filters $g^*_{\phi}(\mathbf{\Lambda})$ will affect the filtered graph signals and produce inaccurate node embeddings.
To measure the changes in spectral filters, we define the spectral distance between the original graph $G$ and perturbed graph $G'$ as:
\begin{equation}
\label{eq:spectral0}
    \mathcal{D}_\text{spectral}=\|g^*_{\phi}(\mathbf{\Lambda})-g^*_{\phi}(\mathbf{\Lambda'})\|_2
\end{equation}
where $\mathbf{\Lambda}$ and $\mathbf{\Lambda}'$ are the eigenvalues of the normalized graph Laplacian for $G$ and $G'$ respectively. 
The spectral distance $\mathcal{D}_\text{spectral}$ is determined by both filter parameters $\phi$ and the frequency components $\mathbf{\Lambda}$. 
For graph embedding models based on the vanilla GCN \cite{kipf2016semi}, we follow their design of spectral filters which
uses the first-order approximation of the Chebyshev polynomials in Eq. (\ref{eq:chev}) and sets $\phi_0=-\phi_1=1$, which gives:
\begin{equation}
\label{eq:gcn-filter}
    g^*_{\phi}(\mathbf{\Lambda})\approx\phi_0\mathbf{I}_0+\phi_1\mathbf{\Lambda}=\mathbf{I}_n-\mathbf{\Lambda}
\end{equation}
Plugging it into Eq. (\ref{eq:spectral0}), we conclude the following spectral distance which is only related to the eigenvalues of graph Laplacian:
\begin{align}
\label{eq:spectral}
    \mathcal{D}_\text{spectral} &\approx \|(\mathbf{I}_n-\mathbf{\Lambda})-(\mathbf{I}_n-\mathbf{\Lambda}')\|_2=\|\mathbf{\Lambda}-\mathbf{\Lambda}'\|_2 \nonumber\\
    &= \|\text{eig}(\text{Laplacian}(\mathbf{A}))-\text{eig}(\text{Laplacian}(\mathbf{A'}))\|_2 
\end{align}
This spectral distance reflects the changes of spectral filters due to the graph perturbation.
Therefore, if we perturb the graph by directly maximizing the spectral distance, we can impose the most effective changes to graph filters and thus disrupt the generated node embeddings the most. 

\subsection{Spectral Attack on Graph Structure}
Since the spectral distance measures the changes of the spectral filters on the graph after perturbation, we can produce effective attacks by maximizing the resulting spectral distance.
In this section, we first show the general formulation of spectral attack on graph structure; then we propose a practical approach to solve the problem efficiently; finally we extend the proposed attack to existing white-box attack frameworks.

\noindent\textbf{Structural perturbation matrix.}
The goal is to find the perturbation on the graph adjacency matrix that can maximize the spectral distance $\mathcal{D}_{\text{spectral}}$ defined in Eq. (\ref{eq:spectral}).
We first define the structural perturbation as a binary perturbation matrix $\mathbf{B}\in\{0, 1\}^{n\times n}$, which indicates where to flip the edges in $G$. The new adjacency matrix after the perturbation is then a function of the perturbation matrix, which can be obtained as follows \cite{xu2019topology}:
\begin{equation}
\label{eq:b}
    g(\mathbf{A, B})=\mathbf{A}+\mathbf{C}\circ\mathbf{B},\ \mathbf{C}=\bar{\mathbf{A}}-\mathbf{A}
\end{equation}
where $\mathbf{\bar{A}}$ is the complement matrix of the adjacency matrix $\mathbf{A}$, calculated by $\mathbf{\bar{A}}=\mathbf{1}_n\mathbf{1}_n^{\top}-\mathbf{I}_n-\mathbf{A}$, with $(\mathbf{1}_n\mathbf{1}_n^{\top}-\mathbf{I}_n)$ denoting the fully-connected graph without self loops.
Therefore, $\mathbf{C}=\bar{\mathbf{A}}-\mathbf{A}$ denotes legitimate addition or deletion operations on each node pair: adding an edge is allowed between node $i$ and $j$ if $\mathbf{C}_{ij}=1$, and removing an edge is allowed if $\mathbf{C}_{ij}=-1$.
Taking the Hadamard product $\mathbf{C}\circ\mathbf{B}$ finally gives valid edge perturbations to the graph.

\noindent\textbf{Spectral attack.}
To generate an effective structural attack, we seek a perturbation matrix $\mathbf{B}$ that maximizes the spectral distance $\mathcal{D}_\text{spectral}$ defined in Eq. (\ref{eq:spectral}). More specifically, given a finite budget of edge perturbation, e.g., $\|\mathbf{B}\|_0\leq \epsilon|E|$ with $|E|$ denoting the number of edges, we formulate the \textbf{SP}ectral \textbf{A}tta\textbf{C}k (\textbf{SPAC}) as the following optimization problem:
\begin{align}
\label{eq:black-box}
    \max_{\mathbf{B}}\  & \mathcal{L}_\text{SPAC} \coloneqq
    \mathcal{D}_\text{spectral} \\
    \text{subject to}\ & \ \|\mathbf{B}\|_0\leq \epsilon|E|, \mathbf{B}\in\{0,1\}^{n\times n}, \mathbf{A'}=g(\mathbf{A, B})\nonumber
\end{align}
which is not straightforward to solve because of two challenges: 1) it is a combinatorial optimization problem due to the binary constraint on $\mathbf{B}$; 2) the objective involves eigen-decomposition of the Laplacian matrix, which is time-consuming especially for large graphs.
Next, we introduce our practical solution to address these challenges such that we can efficiently generate the structural perturbations for the spectral attack.

\subsection{Implementation of \model{}}
In this section, we discuss our solution for the combinatorial optimization problem involving eigen-decomposition in Eq. (\ref{eq:black-box}). Specifically, we first relax the combinatorial problem and use a randomization sampling strategy to generate the binary perturbation matrix; we then introduce an approximation strategy to reduce the complexity of backpropagation through eigen-decompostion.

\noindent\textbf{Binary perturbation via gradient descent}. For the ease of optimization, we relax $\mathbf{B}\in\{0,1\}^{n\times n}$ to its convex hull $\mathbf{\Delta}\in[0,1]^{n\times n}$ \cite{xu2019topology}, which simplifies the combinatorial problem in Eq. (\ref{eq:black-box}) to be the following continuous optimization problem:
\begin{align}
\label{eq:continuous}
    \max_{\mathbf{\Delta}}\ & \mathcal{L}_\text{SPAC} 
    \coloneqq \mathcal{D}_\text{spectral} \\
    \text{subject to}\ & \ \|\mathbf{\Delta}\|_1\leq \epsilon|E|, \mathbf{\Delta}\in[0,1]^{n\times n}, \mathbf{A'}=g(\mathbf{A, \Delta})\nonumber
\end{align}
which can be solved via gradient descent. 
Applying chain rule, we calculate the gradient with respect to $\mathbf{\Delta}$ as follows:
\begin{equation}
\label{eq:gradient}
    \frac{\partial \mathcal{L}_\text{SPAC}}{\partial \mathbf{\Delta}_{ij}}=\sum^n_{k=1}\frac{\partial \mathcal{L}_\text{SPAC}}{\partial \lambda'_k}\sum^{n}_{p=1}\sum^{n}_{q=1}\frac{\partial \lambda'_k}{\partial \mathbf{L}'_{pq}}\frac{\partial \mathbf{L}'_{pq}}{\partial \mathbf{\Delta}_{ij}}
\end{equation}
Recall that $\mathbf{L}'=\text{Laplacian}(\mathbf{A}')$ and $\lambda'_k$ is an eigenvalue of $\mathbf{L}'$ in Eq. (\ref{eq:spectral}).
We now focus on the gradient calculation that involves eigen-decomposition. Since the gradient calculation on the rest is straightforward, we leave it to the appendix.
For a real and symmetric matrix $\mathbf{L}$, one can obtain the derivatives of its eigenvalue $\lambda_k$ and eigenvector $\mathbf{u}_k$ by:
$
    \partial\lambda_k/\partial\mathbf{L}=\mathbf{u}_{k}\mathbf{u}_{k}^{\top},
    \partial\mathbf{u}_k/\partial\mathbf{L}=(\lambda_k\mathbf{I}-\mathbf{L})^{+}\mathbf{u}_k
$
\cite{rogers1970derivatives, magnus1985differentiating}.
Therefore, we can directly obtain the closed-form derivative of the eigenvalues in Eq \eqref{eq:gradient} as:
$\partial\lambda'_k/\partial\mathbf{L}'_{pq}=\mathbf{u}'_{kp}\mathbf{u}'_{kq}$.
Note that the derivative calculation requires distinct eigenvalues. This does not hold for graphs satisfying \textit{automorphism}, which reflects structural symmetry of graphs
\cite{godsil1981full}.
To avoid such cases, we add a small noise term to the adjacency matrix of the perturbed graph\footnote{The form of $(\mathbf{N}+\mathbf{N}^{\top})/2$ is to keep the perturbed adjacency matrix symmetric for undirected graphs.}, e.g., $\mathbf{A}'+ \varepsilon\times(\mathbf{N}+\mathbf{N}^{\top})/2$, where each entry in $\mathbf{N}$ is sampled from a uniform distribution $\mathcal{U}(0,1)$ and $\varepsilon$ is a very small constant.
Such a noise addition will almost surely break the graph automorphism, thus enable a valid gradient calculation of eigenvalues.

Solving the relaxed problem in Eq. (\ref{eq:continuous}) using projected gradient descent gives us a continuous perturbation matrix $\mathbf{\Delta}$ that maximizes the spectral change. To recover valid edge perturbations from the continuous $\mathbf{\Delta}$, we then generate a near-optimal solution for the binary perturbation matrix $\mathbf{B}$ via the randomization sampling strategy \cite{xu2019topology}. Specifically, we use $\mathbf{\Delta}$ as a probabilistic matrix to sample the binary assignments as follows:
\begin{equation}
\label{eq:sample}
    \mathbf{B}_{ij}=
    \begin{cases}
    1, & \text{with probability }\mathbf{\Delta}_{ij} \\
    0, & \text{with probability } 1-\mathbf{\Delta}_{ij}
    \end{cases}
\end{equation}

\noindent\textbf{Time complexity analysis.}
Suppose we take aforementioned projected gradient descent for $T$ steps. For each step, \model{} takes eigen-decomposition with time complexity $\mathcal{O}(n^3)$ and samples binary solution with $\mathcal{O}(n^2)$ edge flips.
The overall time complexity for solving \model{} is $\mathcal{O}(T n^3+T n^2)$, which is mainly attributed to eigen-decomposition and is considerably expensive for large graphs. 
Next, we discuss an approximation solution to improve its efficiency. 

\noindent\textbf{Efficient approximation for \model{}}.
To reduce the computation cost of eigen-decomposition, instead of measuring the spectral distance over all the frequency components, we decide to only maintain the $k_1$ lowest- and $k_2$ highest-frequency components which are the most informative, as suggested by the spectral graph theory (this can also be intuitively observed in Figure \ref{fig:example}). 
Specifically, $\mathcal{D}_\text{spectral}$ in Eq. (\ref{eq:spectral}) can be approximated as follows:
\begin{equation}
\label{eq:approx}
    \mathcal{D}_\text{spectral-approx}=\sqrt{\sum_{i\in\mathcal{S}}(\lambda_i-\lambda'_i)^2}
\end{equation}
where $\mathcal{S}=\{1,\dots,k_1,n-k_2,\dots,n\}$.
This reduces the time complexity $\mathcal{O}(n^3)$ for exact eigen-decomposition to $\mathcal{O}((k_1+k_2)\cdot n^2)$ for the corresponding selective eigen-decomposition using the Lanczos Algorithm \cite{parlett1979lanczos}.
To avoid frequent computation of the selective eigenvalues and further improve efficiency,
we propose to estimate the change in each eigenvalue for any edge perturbation based on the eigenvalue perturbation theory \cite{rellich1969perturbation, bojchevski2019adversarial, chang2020restricted}.

\noindent\textbf{Theorem 1}. \textit{Let $\mathbf{u}_i$ be the $i$-th generalized eigenvector of  $\mathbf{L}$ with generalized eigenvalue $\lambda_i$. Let $\mathbf{L}'=\mathbf{L}+\nabla\mathbf{L}$ be the perturbed Laplacian matrix, and $\mathbf{M}'$ be the diagonal matrix summing over rows of $\mathbf{L}'$. The generalized eigenvalue $\lambda'_i$ of the Laplacian matrix $\mathbf{L'}$ that solves $\mathbf{L}'\mathbf{u}'_i=\lambda'_i\mathbf{M}'\mathbf{u}'_i$ is approximately $\lambda'_i\approx \lambda_i+\nabla\lambda_i$ with:
\begin{align}
\label{eq:theorem}
   \lambda_i-\lambda'_i=\nabla\lambda_i=\mathbf{u}_i^{\top}(\nabla\mathbf{L}-\lambda_i\text{diag}(\nabla\mathbf{L}\cdot\mathbf{1}_n))\mathbf{u}_i
\end{align}}
The proof is given in the appendix. Instead of recalculating the eigenvalues $\lambda'_i$ for the updated $\mathbf{L}'$ in each step when measuring the spectral distance, we use this theorem to approximate the change of each eigenvalue in Eq. (\ref{eq:approx}) in linear time.
Suppose we execute Eq. (\ref{eq:theorem}) to calculate spectral distance in Eq. (\ref{eq:approx}) for $m$ steps and compute the exact eigenvalues every $m$ step to avoid error accumulation, we can achieve an overall time complexity $\mathcal{O}\big((1+\frac{k_1+k_2}{m}) T n^2\big)$. We name the spectral attack equipped with the approximation stated in Eq. (\ref{eq:approx}) and (\ref{eq:theorem}) as \textbf{\model{}-approx}.

Algorithm \ref{alg:spac} summarizes the implementation of \model{} (in line 5) and its approximation \model{}-approx (in line 7-10). After obtaining the objective function based on the (approximated) spectral distance, the algorithm further updates the continuous perturbation matrix $\mathbf{\Delta}$ via gradient descent and finally generates the binary structural perturbation matrix $\mathbf{B}$ by sampling from $\mathbf{\Delta}$.

\subsection{Extension in White-box Setting}
The proposed attack only requires information about graph spectrum, therefore it can be conducted alone in the black-box setting as Eq. (\ref{eq:black-box}) stated. 
Since SPAC does not rely on any specific embedding model, it can also serve as a general recipe for the white-box attack setting. Next, we show how to easily combine SPAC with white-box attack models.

\noindent\textbf{Victim Graph Embedding Model.}
Without loss of generality, we consider the vanilla GCN model for the node classification task.
Given a set of labeled nodes $V_0\subset V$, where each node $i$ belongs to a class in a label set $y_i\in Y$. The GCN model aims to learn a function $f_\theta$ that maps each node to a class.
We consider the commonly studied transductive learning setting, where the test (unlabeled) nodes with associated features and edges are observed in the training phase.
The GCN model is trained by minimizing the following loss function:
\begin{align*}
    \min_{\theta}\  \mathcal{L}_\text{train}(f_\theta(G))=\sum_{v_i\in V_0}\ell(f_{\theta}(\mathbf{A,X})_{i}, y_{i})
\end{align*}
where $f_{\theta}(\mathbf{X,A})_i$ and $y_i$ are the predicted and ground-truth labels of node $v_i$ and $\ell(\cdot,\cdot)$ is a loss function of choice, such as the cross entropy loss.


\begin{algorithm}[tb]
  \caption{Spectral Attack on Graph Structure}
  \label{alg:spac}
  \begin{flushleft}
        \textbf{Input:} $G=(\mathbf{X, A})$; total step $T$; step size $\eta$; $k_1, k_2, m$.
  \end{flushleft}
  \begin{algorithmic}[1]
    \STATE Initialize continuous perturbation $\mathbf{\Delta}_0\in[0,1]^{n\times n}$
    \STATE Initialize perturbed Laplacian matrix $\mathbf{L}'=\text{Laplacian}(g(\mathbf{A},\mathbf{\Delta}))$
    \FOR{$t=0,\dots,T-1$}
        \IF{SPAC}
            \STATE $\mathcal{L}(\mathbf{\Delta})=\|\mathbf{\Lambda}-\mathbf{\Lambda}'\|_2$ by Eq. (\ref{eq:continuous}), with $\mathbf{\Lambda}'=\text{eig}(\mathbf{L}')$
        \ELSIF{SPAC-approx}
            \IF{ $t\ \%\ m = 0$}
                \STATE $\mathcal{L}(\mathbf{\Delta})=\sqrt{\sum_{i\in\mathcal{S}}(\lambda_i-\lambda'_i)^2}$ by Eq. (\ref{eq:approx})
            \ELSE
                \STATE $\mathcal{L}(\mathbf{\Delta})=\sqrt{\sum_{i\in\mathcal{S}}(\mathbf{u}_i^{\top}(\nabla\mathbf{L}-\lambda_i\text{diag}(\nabla\mathbf{L}\cdot\mathbf{1}_n))\mathbf{u}_i)^2}$ by Eq. (\ref{eq:theorem}) and Eq. (\ref{eq:approx}), with $\nabla\mathbf{L}=\mathbf{L'-L}$
            \ENDIF
        \ENDIF
        \STATE Compute gradient on $\mathbf{\Delta}$: $\mathbf{g}_t=\nabla\mathcal{L}(\mathbf{\Delta})$ by Eq. (\ref{eq:gradient})
        \STATE Update $\mathbf{\Delta}_{t+1}=\mathbf{\Delta}_{t}+\eta\cdot\mathbf{g}_t$
        \STATE Project $\mathbf{\Delta}_{t+1}$ to its convex hull $\mathbf{\Delta}_{t+1}\in[0,1]^{n\times n}$
      \ENDFOR
    \STATE Output binary perturbation $\mathbf{B}$ by sampling from $\mathbf{\Delta}_T$ via Eq. (\ref{eq:sample})
  \end{algorithmic}
\end{algorithm}

\noindent\textbf{White-box Spectral Attack.}
We have shown that the changes of the spectral filters are essential for attackers to disrupt graph convolutions, thus we propose to maximize the spectral distance between the original graph and the perturbed one.
In the meanwhile, maximizing the task-specific attack objective is necessary to achieve the attack's goal to compromise the prediction performance. 
To generate edge perturbations that lead to both disrupted spectral filters and erroneous classifications, we propose to maximize the spectral distance and task-specific attack objective simultaneously. 
Specifically, given the test node-set $V_t\subset V$, the attack model aims to find the edge perturbation $\mathbf{B}$ that solves the following optimization problem:
\begin{align}
\label{eq:model}
    \max_{\mathbf{B}}\ & \underbrace{\sum_{v_i\in{\mathcal{V}_t}}\ell_\text{atk}(f_{\theta^{*}}(\mathbf{A',X})_i, y_i)}_\text{task-specific attack objective $\mathcal{L}_{\text{attack}}$} + \beta\cdot\mathcal{L}_\text{SPAC}\nonumber\\
    \text{subject to}\ & \ \|\mathbf{B}\|_0\leq \epsilon, \mathbf{B}\in\{0,1\}^{n\times n} \nonumber, \mathbf{A}'=g(\mathbf{A}, \mathbf{B}) \nonumber\\
    & \ \theta^{*}=\text{arg}\min_{\theta}\ \mathcal{L}_\text{train}(f_{\theta}(\hat{G}))
\end{align}
where the third constraint controls when to apply the attack: setting $\hat{G}=G$ makes it an evasion attack, so that the graph embedding model training is not affected; and setting $\hat{G}=G'$ makes it poisoning attack with perturbed training data.
The attack objective $\mathcal{L}_\text{attack}$ is a flexible placeholder that can adapt to many loss designs, for a simple example, the cross-entropy loss on test nodes in the node classification task.
The hyper-parameter $\beta$ balances the effect of these two components, which is set based on graph properties such as edge density.
Algorithm \ref{alg:spac} also applies for the white-box setting by plugging in $\mathcal{L}_\text{attack}$ to the objective function.
We will discuss the choices of attack objectives and hyper-parameters in the experiment section for our empirical evaluations.


\subsection{Discussion}
Our spectral attack is based on the fact that the spectral filters are the fundamental building blocks for graph convolutions to process graph signals in the Fourier domain.
Therefore, searching the graph perturbations in the direction that causes the most changes in the spectral filters, measured by eigenvalues of graph Laplacian, is expected to best disrupt graph convolutions.
This is also supported by recent theoretical evidence in the field.

Some recent literature has shown that the stability of GCN models is closely related to the eigenvalues of the graph Laplacian. 
For example, it is proved that the generalization gap of a single layer GCN model $f_\theta$ trained via $T$-step SGD is $\mathcal{O}(\lambda_n^{2T})$, where $\lambda_n$ is the largest eigenvalue of graph Laplacian \cite{verma2019stability}.
Meanwhile, \citet{weinberger2007graph} proved that a generalization estimate is \textit{inversely} proportional to the second smallest eigenvalue of the graph Laplacian $\lambda_2$.
These findings suggest that manipulating the graph by perturbing the eigenvalues can potentially aggravate the generalization gap of GCN, causing a larger generalization error. 

\section{Experiments}
We performed extensive evaluations of the proposed spectral attack on four popularly used graph datasets, where we observed remarkable improvements in the attack's effectiveness.
This section summarizes our experiment setup, performance on both evasion and poisoning attacks, and qualitative analysis on the perturbed graphs to study the effect of the spectral attack.

\begin{table}
\caption{Dataset statistics. $D$ is the node feature dimension (``-'' means no node feature). $K$ is the number of classes.}
\vspace{-2mm}
\label{tab:dataset}
\begin{center}
\begin{tabular}[htb]{cccccc}
    \toprule
    Dataset & \#Node & \#Edge & Density & D & K  \\
    \midrule
    Cora & 2,708 & 5,278 & 0.0014 & 1433 & 7 \\
    Citeseer & 3,312 & 4,536 & 0.0008 & 3703 & 6 \\
    Polblogs & 1,490 & 16,715 & 0.015 & - & 2 \\
    Blogcatalog & 5,196 & 171,743 & 0.013 & 8189 & 6 \\
    \bottomrule
\end{tabular}
\end{center}
\vspace{-3mm}
\end{table}

\begin{table}
\caption{Average running time (in seconds) for $10$ runs of evasion attack with $T=100$ and $\epsilon=0.05$.}
\small
\label{tab:time}
\vspace{-2mm}
\begin{center}
\begin{tabular}[htb]{cccccc}
    \toprule
    Datasets & Random & DICE & GF-Attack & \textbf{SPAC} & \textbf{SPAC-approx} \\
    \midrule
    Cora & 0.05 & 55.58 & 66.73 & 212.53 & 75.46 \\
    Citeseer & 0.06 & 46.72 & 57.22& 116.07 & 60.93 \\
    Polblogs & 0.02 & 14.84 & 21.73 & 44.18 & 22.98 \\
    Blogcatalog & 1.46 & 127.72 & 132.23 & 352.52 & 147.34 \\
    \bottomrule
\end{tabular}
\end{center}
\vspace{-3mm}
\end{table}

\begin{table}[tb]
\caption{Misclassification rate (\%) with $\epsilon=0.05$ for evasion attack (upper rows) and poisoning attack (lower rows).}
\vspace{-2mm}
\label{tab:result}
\begin{center}
\small
\setlength{\tabcolsep}{0.5em}
\begin{tabular}[htb]{c|ccccc}
\toprule
    Stage & Attack &  Cora & Citeseer & Polblogs & Blogcatalog \\
    \midrule
    \multirow{10}{*}{Evasion} & 
    Clean & 0.184 & 0.295 & 0.128 & 0.276 \\
     & Random & 0.189 & 0.301 & 0.153 & 0.280 \\
     & DICE & 0.205 & 0.308 & 0.202 & 0.329 \\
     & GF-Attack & 0.198 & 0.311 & 0.179 & 0.333\\
     & \textbf{SPAC} & \textbf{0.220} & \textbf{0.314} & \textbf{0.212} & \textbf{0.354} \\
     & \textbf{SPAC-approx} & 0.212 & 0.305 & 0.208 & 0.341\\
    \cmidrule{2-6}
     & PGD-CE & 0.237 & 0.349 & 0.167 & 0.441 \\
     & \textbf{\model{}-CE} & \textbf{0.255} & \textbf{0.352} & \textbf{0.188} & \textbf{0.458} \\
    \cmidrule{2-6}
     & PGD-C\&W & 0.249 & 0.388 & 0.216 & 0.447\\
     & \textbf{\model{}-C\&W} & \textbf{0.260} & \textbf{0.395} & \textbf{0.229} & \textbf{0.464} \\
    \bottomrule
    \multirow{12}{*}{Poison} & Clean & 0.184 & 0.295 & 0.128 & 0.276 \\
     & Random & 0.189 & 0.309 & 0.126 & 0.277 \\
     & DICE & 0.207 & 0.310 & \textbf{0.246} & 0.306\\
     & GF-Attack & 0.195 & 0.306 & 0.202 & 0.334\\
     & \textbf{SPAC} & \textbf{0.222} & \textbf{0.338} & 0.234 & \textbf{0.478} \\
     & \textbf{SPAC-approx} & 0.215 & 0.322 & 0.220 & 0.454\\
    \cmidrule{2-6}
     & Max-Min & 0.240 & 0.359 & 0.167 & 0.489 \\
     & \textbf{\model{}-Min} & \textbf{0.255} & \textbf{0.375} & \textbf{0.188} & \textbf{0.504} \\
    \cmidrule{2-6}
     & Meta-Train & \textbf{0.290} & 0.392 & 0.274 & 0.360\\
     & \textbf{\model{}-Train} & 0.285 & \textbf{0.412} & \textbf{0.298} & \textbf{0.377} \\
    \cmidrule{2-6}
     & Meta-Self & 0.427 & 0.499 & \textbf{0.478} & 0.590\\
     & \textbf{\model{}-Self} & \textbf{0.489} & \textbf{0.508} & 0.472 & \textbf{0.599}\\
    \bottomrule
\end{tabular}
\end{center}
\vspace{-3mm}
\end{table}


\subsection{Setup}
\textbf{Datasets}. 
We evaluated the proposed attack on two citation network benchmark datasets, Cora \cite{mccallum2000automating} and Citeseer \cite{sen2008collective}, as well as two social network datasets, Polblogs \cite{adamic2005political} and Blogcatalog \cite{rossi2015network}. 
Table \ref{tab:dataset} summarizes the statistics of these datasets. 
We followed the transductive semi-supervised node classification setup in \cite{kipf2016semi}, where only $20$ sampled nodes per class were used for training, but the features and edges of all nodes were visible to the attacker during the training stage. 
The predictive accuracy of the trained classifier was evaluated on $1000$ randomly selected test nodes.
The evasion attacker can query the trained classifier, but cannot access the training nodes; the poisoning attacker can observe the training nodes, but cannot access the labels of the test nodes.

\begin{figure*}[htb]
 \centering
  \includegraphics[width=1.0\textwidth]{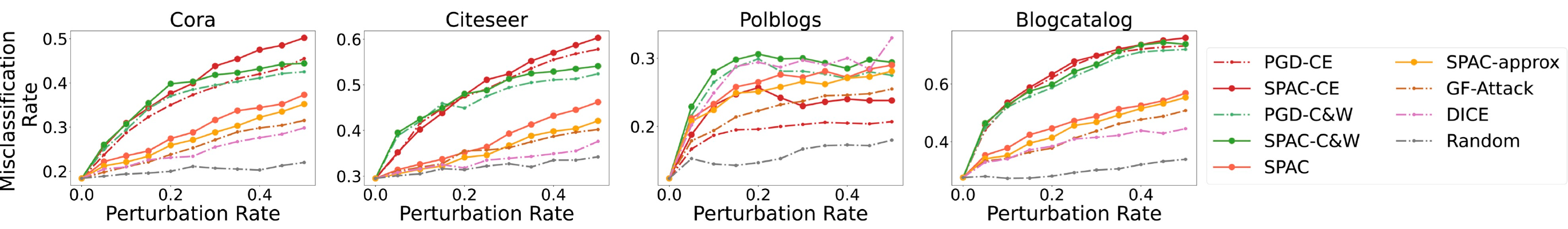}
  \vspace{-4mm}
 \caption{Misclassification rate under different perturbation rates for evasion attack.}
 \vspace{-4mm}
 \label{fig:evasion}
\end{figure*}

\begin{figure*}[htb]
 \centering
  \includegraphics[width=1.0\textwidth]{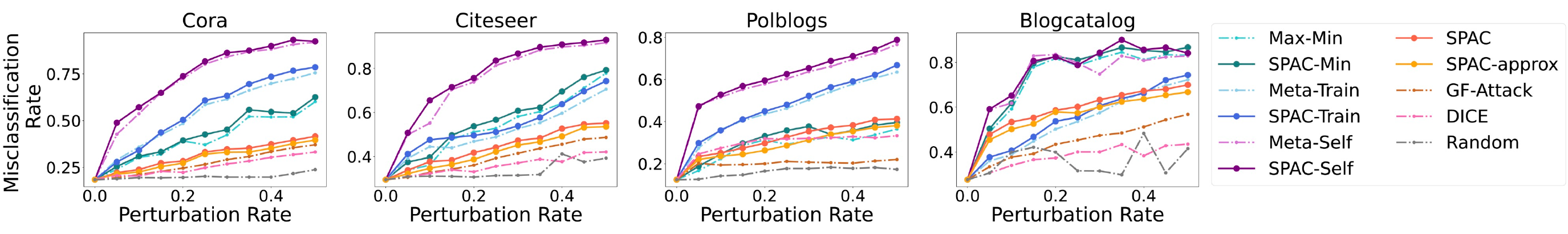}
  \vspace{-3mm}
 \caption{Misclassification rate under different perturbation rates for poisoning attack.}
 \vspace{-3mm}
 \label{fig:poison}
\end{figure*}

\noindent\textbf{Baseline attacks}. 
We compared the proposed attack model \model{} against three attacks in the black-box setting, and further verified the effectiveness of \model{} by combining it with five baselines in white-box setting\footnote{We conducted the comparative experiments using DeepRobust Library \cite{li2020deeprobust}.}. 
Black-box baselines for both \textit{evasion} and \textit{poisoning} attack include:  
1) \textbf{Random} directly attacks the graph structure by randomly flipping the edges;
2) \textbf{DICE} \citep{waniek2018hiding} is a heuristic method that deletes edges internally and connects nodes externally across class clusters;
3) \textbf{GF-Attack} \cite{chang2020restricted} perturbs the structure by maximizing the loss of low-rank matrix approximation defined over small eigenvalues.
We further evaluate the performance of \model{} combined with white-box attack baselines which include:
1) \textbf{PGD-CE} \citep{xu2019topology} is an \textit{evasion} attack which maximizes the cross-entropy (CE) loss on the test nodes via projected gradient descent (PGD) algorithm \citep{madry2017towards};
2) \textbf{PGD-C\&W} \citep{xu2019topology} is an \textit{evasion} attack which perturbs edges by minimizing the C\&W score, which is the margin between the largest and the second-largest prediction score, defined by Carlini-Wagner attack \citep{carlini2017towards};
3) \textbf{Max-Min} \citep{xu2019topology} is a \textit{poisoning} attack, which solves the bi-level optimization problem by iteratively generating structural perturbations (to maximize the cross-entropy loss) and retraining a surrogate victim model on the perturbed graph (to minimize the loss);
4) \textbf{Meta-Train} \citep{zugner2019adversarial} is a \textit{poisoning} attack which uses meta-gradients on the perturbation matrix to maximize the training classification loss;
5) \textbf{Meta-Self} \citep{zugner2019adversarial}  is a \textit{poisoning} attack that extends Meta-Train to maximize the self-training loss on test nodes using the predicted labels;

\noindent\textbf{Variants of \model{}.} 
The proposed attack can be realized with exact and approximated spectral distance, which gives \textbf{\model{}} and \textbf{\model{}-approx}. We will compare their attack performance and running time.
Meanwhile, adopting the objectives from white-box baselines to Eq. (\ref{eq:model}) generates the following white-box attack variants:
1) \textbf{\model{}-CE} is an \textit{evasion} attack that jointly maximizes the cross-entropy loss and spectral distance;
2) \textbf{\model{}-C\&W} is an \textit{evasion} attack combining the \textit{negative} C\&W score and spectral distance;
3) \textbf{\model{}-Min} extends Max-Min by maximizing the loss as in \model{}-CE for poisoning attack;
4) \textbf{\model{}-Train} includes the spectral distance to Meta-Train for the meta-gradient calculation;
5) \textbf{\model{}-Self} enhances the loss of Meta-Train by the spectral distance.
The detailed objective for each variant is summarized in the appendix.

\noindent\textbf{Hyper-parameters.}
We adopt a two-layer GCN as the victim classifier, whose hidden dimension of the first layer is $64$ and that of the second layer is the number of classes $K$ on each dataset. 
The setup for total step $T$ and step size $\eta$ is summarized in the appendix.
For \model{}-approx, we set $k_1=128$, $k_2=64$ and $m=10$.
In the white-box setting, the hyper-parameter $\beta$ controls the weight of the spectral distance in the overall attack objective.
Since the spectrum reflects the global property of the graph, the weight should be tuned based on the graph statistics.
Empirically, we find that setting $\beta$ proportional to the density of graph is effective. Specifically, according to the density of each dataset listed in Table \ref{tab:dataset}, we set $\beta=1.4$ for Cora network, $\beta=0.8$ for Citeseer, $\beta=13.0$ for Blogcatalog, and $\beta=15.0$ for Polblogs.
The sensitivity of hyper-parameters is discussed in Section \ref{sec:analysis}.
All experiments were conducted on RTX2080Ti GPUs.

\subsection{Structural Attack Performance}

\noindent\textbf{Performance in evasion attack.}
In the evasion attack setting, we first trained a GCN classifier on the small training set $V_0$ with a clean graph $\mathcal{\hat{G}}=\mathcal{G}$. Then the classifier was fixed, and the attackers generated edge perturbations based on the classifier's predictions on the test nodes.
Table \ref{tab:result} summarizes the misclassification rates under $\epsilon=0.05$, which allows $5\%$ edges to be perturbed.
An extensive comparison with different perturbation rates is provided in Figure \ref{fig:evasion}, where the solid lines with darker color denote \model{} variants while the dashed lines with lighter color represent baseline attacks.

In the black-box setting, randomly flipping edges (Random) cannot effectively influence the classifier's overall performance.
DICE provides an effective attack by leveraging the label information.
GF-Attack undermines the performance of GCNs by attacking its low-rank approximation.
Our methods, both \model{} and \model{}-approx, disrupt the overall spectral filters and achieve the largest misclassification rate.
This shows the effectiveness of the proposed attack principle based on the spectral distance, which reveals the essence of vulnerability in graph convolutions. 

Second in the white-box setting, \model{}-CE and \model{}-C\&W stand in stark contrast to PGD-CE and PGD-C\&W: we can observe a remarkable improvement introduced by SPAC in the misclassification rate.
The evasion attack results confirm that maximizing the spectral distance can considerably disrupt the trained classifier by changing the graph frequencies in the Fourier domain and invalidating the spectral filters.

\noindent\textbf{Performance in poisoning attack.}
In the poisoning attack setting, we can only indirectly affect the classifier by perturbing the training graph structure. We generated the edge perturbations, and then used the poisoned structure to train the victim GCN model and reported its misclassification rate on test nodes in a clean graph. 
From Table \ref{tab:result} and Figure \ref{fig:poison}, we can again verify the effectiveness of the proposed spectral attack.
Under the black-box setting, \model{} and \model{}-approx are the most effective attacks in most cases.
Under the white-box setting, 
Max-Min only accesses training nodes to perturb the graph without querying test nodes.
Meta-Train calculates the meta-gradient on training nodes to capture the change of loss after retraining the surrogate GCN model.
Meta-Self instead does not use the training nodes, but only queries CGN's prediction scores on test nodes.
Among baselines, the Meta-Self attack is shown to be the most effective, which is expected, because the current semi-supervised setting provides a much larger set of unlabeled nodes that can be fully used by Meta-Self. 
Overall, our attack based on the spectral distance still brought in a clear improvement to the misclassification rate across different datasets and attack methods.

\noindent\textbf{Computational efficiency.}
We empirically evaluated the efficiency of \model{} and \model{}-approx in Table \ref{tab:time}, which compares the average running time of $10$ runs for evasion attack.
Our proposed \model{}-approx can achieve a comparable efficiency as GF-Attack.
Combining with the attack performance, \model{}-approx is verified to be an effective and efficient structural attack.

\begin{figure}[ht]
 \centering
  \includegraphics[width=0.4\textwidth]{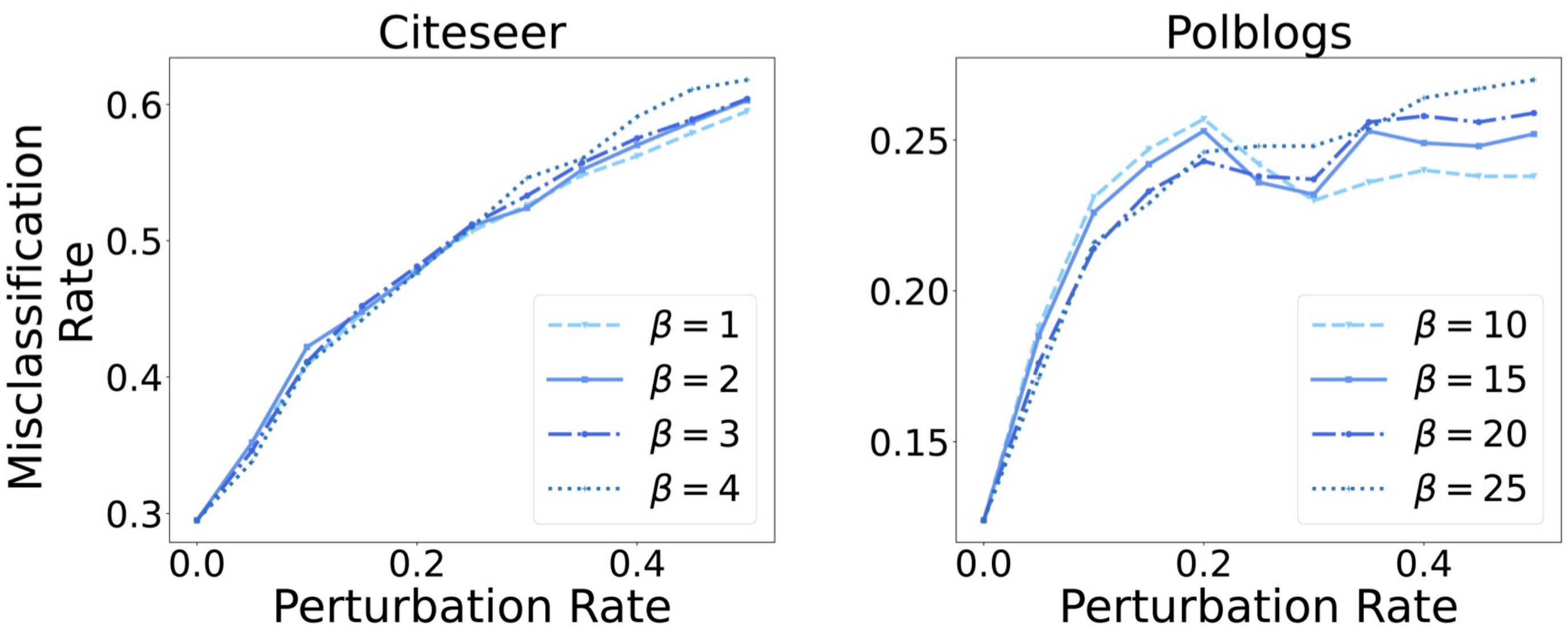}
  \vspace{-1mm}
 \caption{Sensitivity analysis on $\beta$ under \model{}-CE attack.}
 \vspace{-1mm}
 \label{fig:sensitivity_b}
\end{figure}

\begin{figure}[ht]
 \centering
  \includegraphics[width=0.42\textwidth]{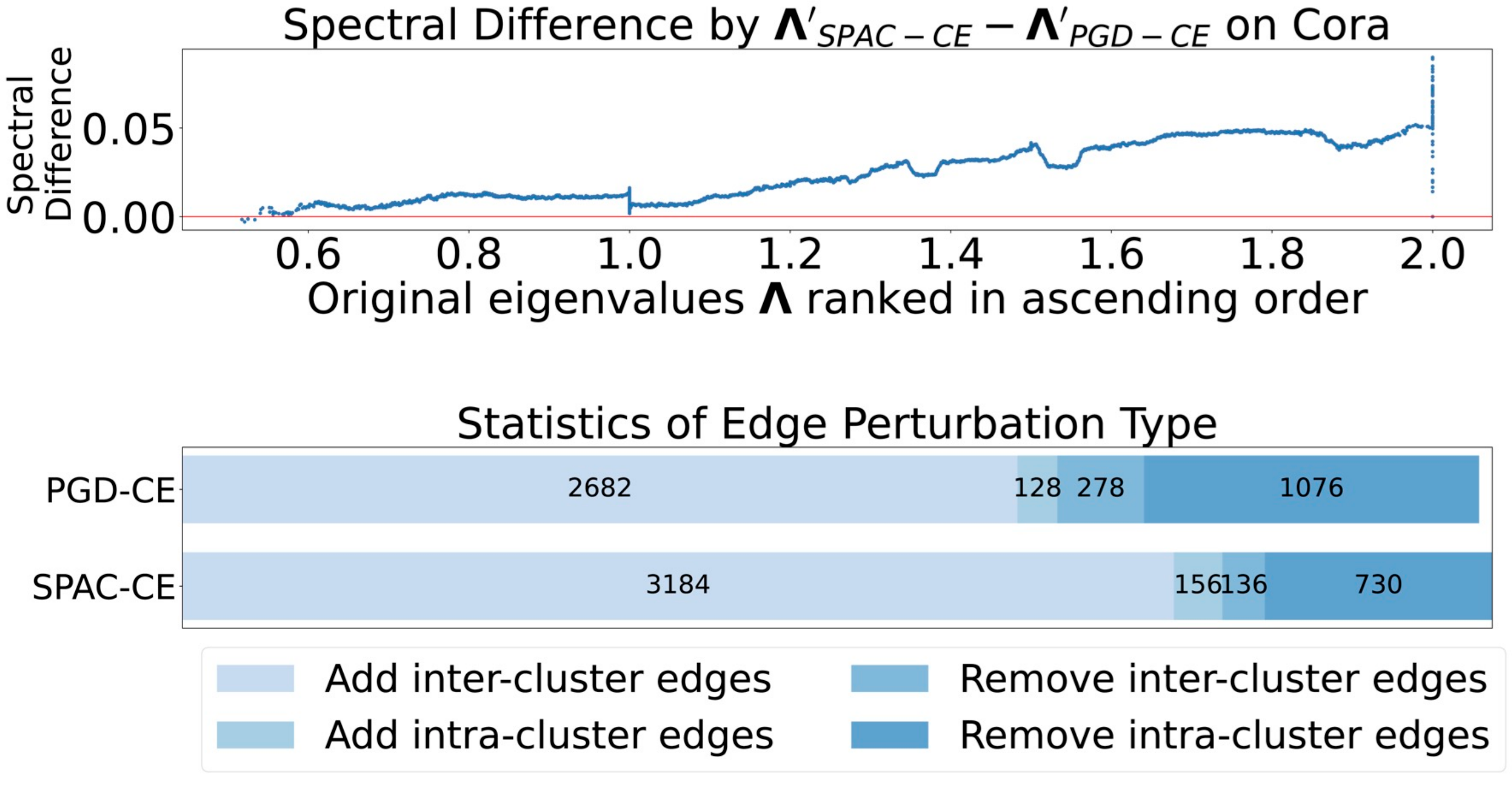}
 \caption{Analysis on  spectral changes (top) and spatial edge changes (bottom) between SPAC-CE and PGD-CE.}
 \label{fig:case}
 \vspace{-2mm}
\end{figure}

\vspace{-2mm}
\subsection{Analysis of \model{}}
\label{sec:analysis}
Given the empirical effectiveness of the proposed attack strategy, we now analyze the sensitivity of hyper-parameters including $k_1,k_2$ and $m$ for \model{}-approx and $\beta$ for white-box setting. We also illustrate the behavior of \model{} in both Fourier and spatial domains.

\begin{figure}[tb]
 \centering
  \includegraphics[width=0.45\textwidth]{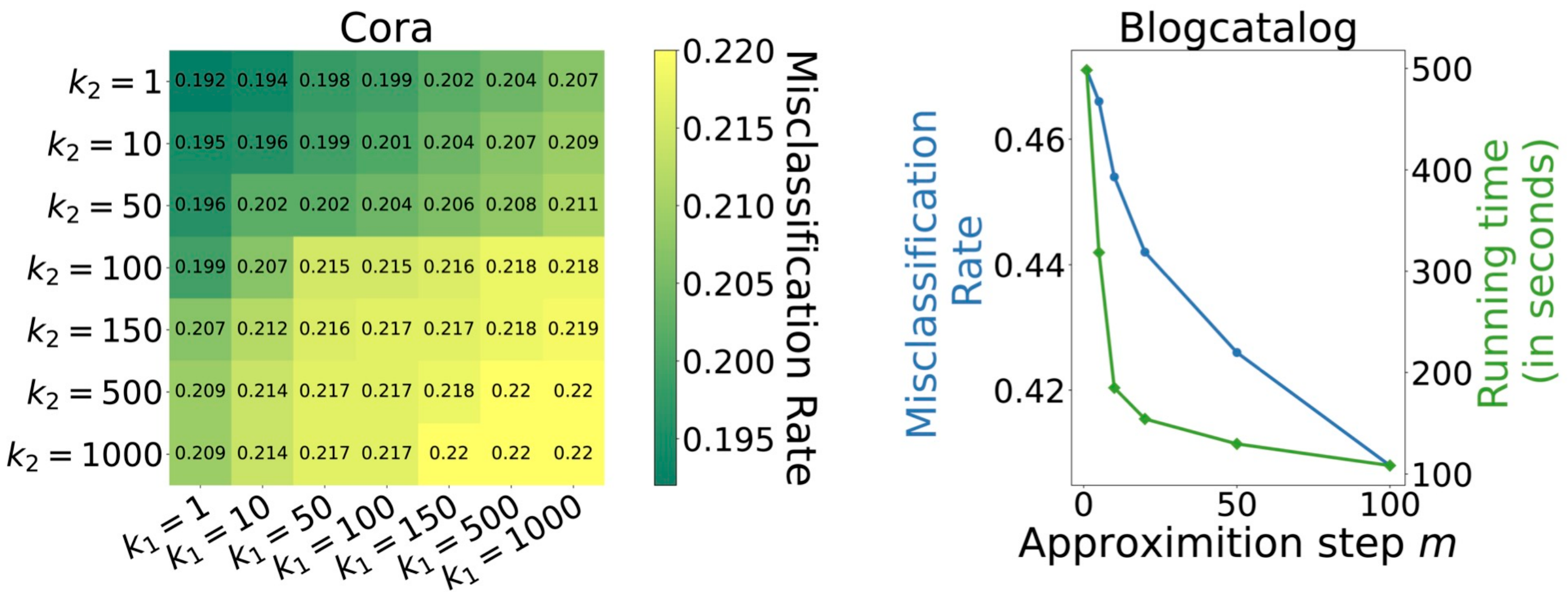}
 \caption{Sensitivity analysis on $k_1, k_2$ (left) and $m$ (right) under \model{}-approx poisoning attack.}
 \label{fig:sensitivity_approx}
 \vspace{-3mm}
\end{figure}

\noindent\textbf{Sensitivity of $k_1, k_2$ and $m$ in \model{}-approx}.
For \model{}-approx, the trade-off of its attack performance and efficiency is achieved by selecting $k_1$ low- and $k_2$ high-frequency components and by approximating the eigenvalue change for $m$ steps.
Figure \ref{fig:sensitivity_approx} demonstrates such trade-off under \model{}-approx poisoning attack with budget $\epsilon=0.05$. Left side shows the misclassification rate range when using different $k_1$ and $k_2$; right side compares the misclassification rate and running time when using different approximation step $m$.
The result suggests that the attack performance does not dramatically drop with changed parameters, and we can achieve a good balance between attack effectiveness and efficiency. 

\noindent\textbf{Sensitivity of hyper-parameter $\beta$ in white-box setting}. 
Figure \ref{fig:sensitivity_b} shows the performance of \model{}-CE under different settings of the coefficient parameter $\beta$.
We can clearly observe that different $\beta$ values lead to rather stable performance, which suggests the spectral distance term can be applied to real applications without the requirement of tedious hyper-parameter tuning. 

\begin{figure}[tb]
 \centering
  \includegraphics[width=0.4\textwidth]{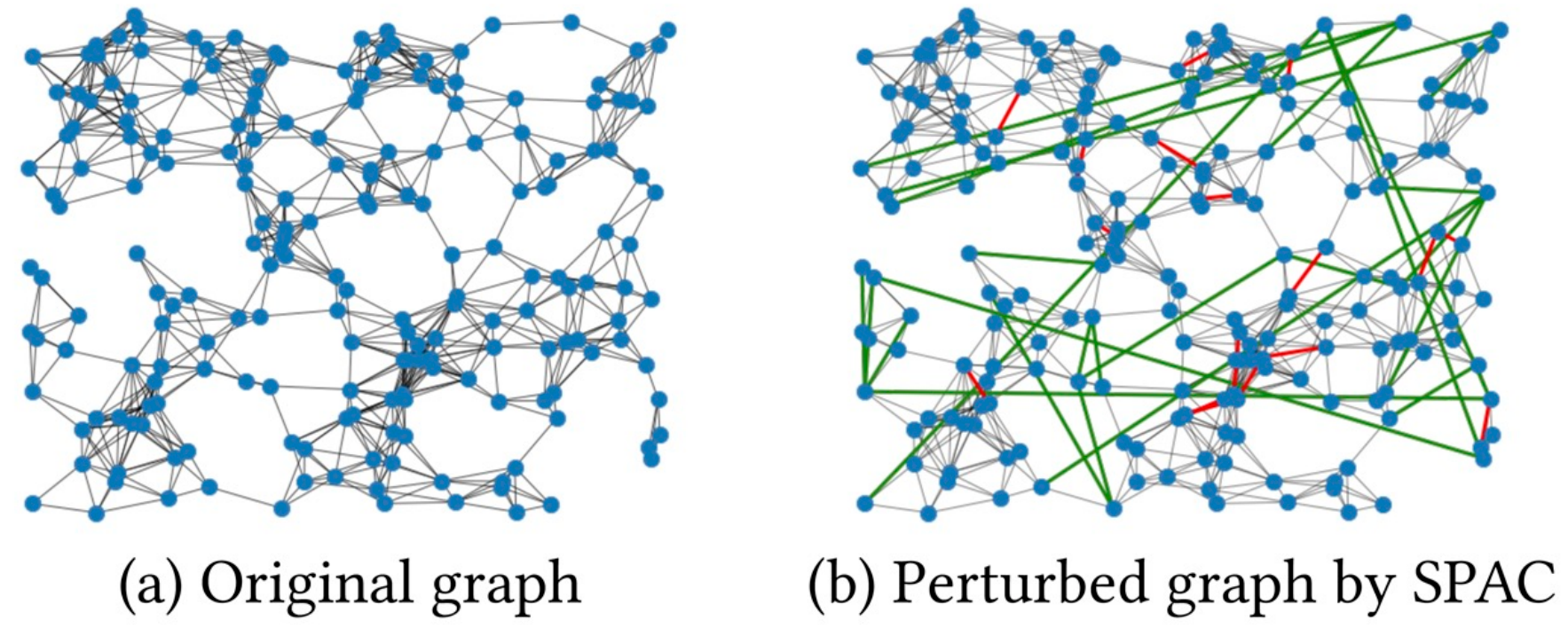}
 \vspace{-3mm}
 \caption{The edge perturbation generated by \model{} on a random geometric graph with $\epsilon=0.05$.  Green denotes edges added by the attack, while red marks removed edges.}
 \label{fig:case2}
 \vspace{-4mm}
\end{figure}

\noindent\textbf{Effect of \model{} in Fourier and spatial domain}.
We are interested in investigating how the changes of the graph in the Fourier domain affect its spatial structure. To serve this purpose, we compared the output of the perturbed graphs from \model{}-CE and PGD-CE on Cora under budget $\epsilon=0.4$ in Figure \ref{fig:case}.
The top plots the difference between eigenvalues of the normalized Laplacian matrix for the graph perturbed by \model{-CE} and the graph perturbed by PGD-CE. 
The x-axis shows the eigenvalues of the original graph.
The bottom counts the number of different types of edge perturbations, where ``inter-cluster'' edges are those connecting nodes with different class labels and ``intra-cluster'' edges connect nodes with the same class label. 
We observe that \model{}-CE perturbed graph in a direction leading to larger high eigenvalues and smaller low eigenvalues, compared with PGD-CE. 
This spectral difference in the Fourier domain is also reflected in the spatial domain: 1) more edges are added than removed; 2) specifically, more inter-cluster edges were added while fewer inter-cluster edges were removed.
To intuitively demonstrate the perturbations generated by \model{}, we applied \model{} to attack the random geometric graph in Figure \ref{fig:example} with budget $\epsilon=0.05$, and the perturbed graph is visualized in Figure \ref{fig:case2}. The green edges that are added by \model{} connect different node clusters, while red edges are removed within clusters.
This shows that maximizing the spectral distance can modify the global connectivity of the graph: for example, \model{}-CE strengthened the connectivity between different clusters to confuse the classifier.

\section{Conclusion}
In this paper, we propose a novel graph structural attack strategy by maximizing the spectral distance between the original and the perturbed graphs. The design is motivated by the spectral perspective for understanding GCNs.
An efficient approximation solution is further designed to reduce the computational complexity of spectral distance.
Our experiments demonstrated the effectiveness of this new direction, and our qualitative study suggested that the proposed spectral attack tends to modify the global connectivity of a graph and enlarge the generalization gap of GCN models.

Currently, we focused on perturbing the eigenvalues of graph Laplacian, without controlling the eigenvectors. As the eigenvectors also play an important role in the spectral filters, it is important to expand our scope to manipulate eigenvectors for improved effectiveness.
Applying the model-agnostic \model\ to attack a broader group of graph embedding models will also be interesting to understand the fundamental vulnerability of the graph structure.

\begin{acks}
This work is supported by the National Science Foundation under grant IIS-1718216, IIS-2128019, and IIS-1553568.
\end{acks}

\bibliographystyle{ACM-Reference-Format}
\balance
\bibliography{5-reference}

\clearpage
\section{Appendix}
We list the detailed gradient calculation of the spectral distance term with eigen-decomposition, the proof of Theorem 1, the attack objectives for different white-box variants of \model{} and the hyper-parameter setup.

\subsection{Gradient of the Spectral Distance}
Recall that we obtain the following form via the chain rule:
\begin{equation*}
    \frac{\partial \mathcal{L}_\text{SPAC}}{\partial \mathbf{\Delta}_{ij}}=\sum^n_{k=1}\frac{\partial \mathcal{L}_\text{SPAC}}{\partial \lambda'_k}\sum^{n}_{p=1}\sum^{n}_{q=1}\frac{\partial \lambda'_k}{\partial \mathbf{L}'_{pq}}\frac{\partial \mathbf{L}'_{pq}}{\partial \mathbf{\Delta}_{ij}}
\end{equation*}
Here is the detailed calculation of each component:
\begin{align*}
    & \frac{\partial \mathcal{L}_\text{SPAC}}{\partial \lambda'_k} = \frac{\lambda'_k-\lambda_k}{\|\mathbf{\Lambda}-\mathbf{\Lambda}'\|_2}\\
    &\frac{\partial \lambda'_k}{\partial \mathbf{L}'_{pq}} = \mathbf{u}'_{kp}\mathbf{u}'_{kq}\\
    &\frac{\partial \mathbf{L}'_{pq}}{\partial \mathbf{\Delta}_{ij}} = \frac{\mathbf{C}_{ij}}{2\sqrt{d'_p d'_q}}(\mathbf{1}_{i=p}\frac{\mathbf{A}'_{pq}}{d'_p}+\mathbf{1}_{j=q}\frac{\mathbf{A}'_{pq}}{d'_p}-2\mathbf{1}_{i=p, j=q})\\
\end{align*}
where $d'_p$ is the degree on node $p$ of the perturbed graph: $d'_p=\sum_{k=1}^{n}\mathbf{A}'_{kp}$, and similarly $d'_q=\sum_{k=1}^{n}\mathbf{A}'_{kq}$. Meanwhile, $\mathbf{A}'$ is the adjacency matrix of the perturbed graph. The indication function $\mathbf{1}_{\text{condition}}$ is $1$ if the condition is true, otherwise it is $0$.

\subsection{Proof of Theorem 1}
\noindent\textit{Proof.} \textbf{Theorem 1.} For the generalized eigenvalue problem: $\mathbf{L}\mathbf{u}_i=\lambda_i\mathbf{M}\mathbf{u}_i$, if the matrix is slightly perturbed $\mathbf{L}'=\mathbf{L}+\nabla \mathbf{L}$, we aim to find the corresponding eigenvalue perturbation: $\lambda'_i=\lambda_i+\nabla \lambda_i$.
From eigenvalue perturbation theory \cite{stewart1990matrix}, we have 
$$\mathbf{\lambda'_i-\lambda_i\approx \mathbf{u}^\top_i(\nabla \mathbf{L}-\lambda_i\nabla \mathbf{M})\mathbf{u}_i}$$
And for a normalized graph Laplacian $\mathbf{L}'=\mathbf{L}+\nabla\mathbf{L}$, we have $\nabla\mathbf{M}=\text{diag}(\nabla \mathbf{L}\cdot\mathbf{1}_n)$. 
Submitting $\nabla \mathbf{M}$ concludes the proof.

\subsection{Attack Objectives for White-box Variants}
Recall that \model{} can be flexibly combined with the white-box attack framework as shown in Eq. (\ref{eq:model}), which consists of a task-specific attack objective $\mathcal{L}_\text{attack}$ and the proposed \model{} objective $\mathcal{L}_\text{SPAC}$. 
We denote the training node set as $V_0$ and test node set as $V_t$. Different choices of $\mathcal{L}_\text{attack}$ result in the following variants.

\noindent\textbf{\model{}-CE} combines \model{} with PGD-CE \cite{xu2019topology}, and maximizes the \textit{cross-entropy} loss on the target \textit{test set} for \textit{evasion} attack:
\begin{equation*}
    \mathcal{L}_\text{attack}=\sum_{v_i \in V_t}\text{crossEntropy}(f_{\theta}(\mathbf{A+\Delta, X})_i, y_i)
\end{equation*}

\noindent\textbf{\model{}-C\&W} combines \model{} wih PGD-C\&W \cite{xu2019topology}, and maximizes the \textit{negative} C\&W \textit{score} on the target \text{test set} for \textit{evasion} attack:
\begin{align*}
    \mathcal{L}_\text{attack}=-\sum_{v_i \in V_t} \text{max}\{Z_{i, y_i}-\max_{c\neq y_i}Z_{i,c}-\kappa\}
\end{align*}
where $Z_{i,c}$ denotes the prediction logit on label $c$, and $\kappa\geq 0$ is a confidence level of making wrong decisions.
Intuitively, the C\&W score evaluates how good the model can differentiate the prediction on the ground-truth label and on the label with the (second) highest likelihood. So the attack aims to confuse the model by maximizing the negative C\&W score.

\noindent\textbf{\model{}-Min} combines \model{} and Max-Min \cite{xu2019topology}, and maximizes the \textit{cross-entropy} loss on the \textit{training set}, while a surrogate model $f_{\theta'}$ is iteratively retrained. The perturbed graph is then used to train a victim model, and we report the classification performance of the test set on clean graph. The $poisoned$ graph is generated by:
\begin{equation*}
    \mathcal{L}_\text{attack}=\sum_{v_i \in V_0}\text{crossEntropy}(f_{\theta'}(\mathbf{A+\Delta, X})_i, y_i)
\end{equation*}

\noindent\textbf{\model{}-Train} combines \model{} with Meta-Train \cite{zugner2019adversarial}, and maximizes the \textit{cross-entropy} loss on labeled \textit{training nodes}, arguing that if a model has a high training error, it is likely to generalize poorly:
\begin{equation*}
    \mathcal{L}_\text{attack}=\sum_{v_i\in V_0}\text{crossEntropy}(f_{\theta'}(\mathbf{A+\Delta, X})_i, y_i)
\end{equation*}
The objective is similar to \model{}-Min, but instead of retraining the surrogate model, \model{}-Train calculate \textit{meta-gradients} on the perturbation matrix through the surrogate model.

\noindent\textbf{\model{}-Self} combines \model{} with Meta-Self \cite{zugner2019adversarial}, and maximizes the \textit{cross-entropy} loss on unlabeled \textit{test nodes} which are assigned pseudo labels predicted by the model trained on tbe clean graph:
\begin{equation*}
    \mathcal{L}_\text{attack}=\sum_{v_i\in V_t}\text{crossEntropy}(f_{\theta'}(\mathbf{A+\Delta, X})_i, \hat{y}_i)
\end{equation*}
where $\hat{y}_i$ is the predicted label from the model trained on the clean graph $f_\theta$.

\subsection{Hyper-parameter Setup}
\noindent For attack methods that involve projected gradient descend (e.g., \model{}/PGD-CE, \model{}/PGD-C\&W, \model{}/PGD-Min, \model{} and \model{}-approx), we optimize the attack objective by gradient descent for $T=100$ iterations; we set adaptive step size for gradient descent as $\eta=T\cdot\epsilon/\sqrt{t}$, which is related to the perturbation budget ratio $\epsilon$, such that for each step we can use up the budget while not exceeding the budget too much.
For \model{}/Meta-Train and \model{}/Meta-Self, the iteration is decided by the perturbation budget: for each step, choose the edge entry that has the largest gradient.
For GF-Attack, the top-$K$ smallest eigenvalues are selected for $K$-rank approximation with $K=n-128$ following the paper's setting. For our approximation model \model{}-approx, the reported results are based on $k_1=128$ lowest eigenvalues and $k_2=64$ largest eigenvalues.

\end{document}